\documentclass[runningheads]{llncs}
\usepackage{graphicx}
\usepackage{amsmath,amssymb} 
\usepackage{epstopdf}
\usepackage{color}

\usepackage{cite}
\usepackage{subfigure}
\usepackage{multirow}
\usepackage{diagbox}
\usepackage{rotating}
\usepackage{booktabs}
\usepackage{colortbl}
\usepackage{overpic}
\usepackage[table]{xcolor}
\usepackage{textcomp}
\usepackage{contour}
\usepackage{textcomp}

\definecolor{mygreen}{RGB}{0,150,0}
\definecolor{myred}{RGB}{200,0,0}
\usepackage[pagebackref=false,letterpaper=true,colorlinks=true,linkcolor=myred,citecolor=blue,bookmarks=false]{hyperref}

\newcommand{\cmark}{\checkmark}
\newcommand{\tickNo}{\texttimes}

\newcommand{\figref}[1]{Fig. \ref{#1}}
\newcommand{\tabref}[1]{Table \ref{#1}}

\newcommand{\secref}[1]{Sec. \ref{#1}}

\renewcommand{\tabcolsep}{.5mm}

\newcommand{\sArt}{state-of-the-art}
\newcommand{\eg}{\emph{e.g.}}
\newcommand{\ie}{\emph{i.e.}}
\newcommand{\etal}{\emph{et al.}}
\newcommand{\AddText}[3]{\put(#1,#2){\contour{white}{\textbf{\textcolor{black}{#3}}}}}
\newcommand{\AddAttr}[3]{\put(#1,#2){\contour{black}{\textbf{\textcolor{white}{#3}}}}}
\graphicspath{{./Imgs/}}

\begin{document}
\title{Salient Objects in Clutter: Bringing Salient Object Detection to the Foreground} 

\titlerunning{Bringing Salient Object Detection to the Foreground}
%
\author{Deng-Ping Fan\inst{1}\orcidID{0000-0002-5245-7518} \and
Ming-Ming Cheng\thanks{ \scriptsize {M.M. Cheng (cmm@nankai.edu.cn) is the corresponding author.
This research was supported by NSFC (NO. 61620106008, 61572264),
the national youth talent support program,
Tianjin Natural Science Foundation for Distinguished Young Scholars (NO. 17JCJQJC43700),
Huawei Innovation Research Program.
}}\inst{,1}\orcidID{0000-0001-5550-8758} \and
Jiang-Jiang Liu\inst{1}\orcidID{0000-0002-1341-2763} \and
Shang-Hua Gao\inst{1}\orcidID{0000-0002-7055-2703} \and
Qibin Hou\inst{1}\orcidID{0000-0002-8388-8708} \and 
Ali Borji\inst{2}\orcidID{0000-0001-8198-0335}}
%
\authorrunning{Fan \etal}
%

\institute{CCCE, Nankai University, Tianjin, China \and
CRCV, University of Central Florida, Orlando, Florida, U.S \\
\url{http://mmcheng.net/SOCBenchmark/}}
\maketitle              
\begin{abstract}
   We provide a comprehensive evaluation of salient object detection (SOD) models.
   Our analysis identifies a serious design bias of existing SOD datasets which
   assumes that each image contains at least one clearly outstanding salient
   object in low clutter.
   The design bias has led to a saturated high performance for \sArt~SOD models when evaluated on
   existing datasets. The models, however, still perform far from being satisfactory when applied
   to real-world daily scenes.
   Based on our analyses, we first identify 7 crucial aspects that a comprehensive and
   balanced dataset should fulfill. Then, we propose a new high quality dataset and update the
   previous saliency benchmark. Specifically, our \textbf{SOC} (Salient Objects
   in Clutter) dataset, includes images with salient and non-salient objects from daily object categories.
   Beyond object category annotations, each salient image is
   accompanied by attributes that reflect common challenges in real-world scenes.
   Finally, we report attribute-based performance assessment on our dataset.

\end{abstract}
\begin{figure}[t!]
	\centering
	\begin{overpic}[width=.75\columnwidth]{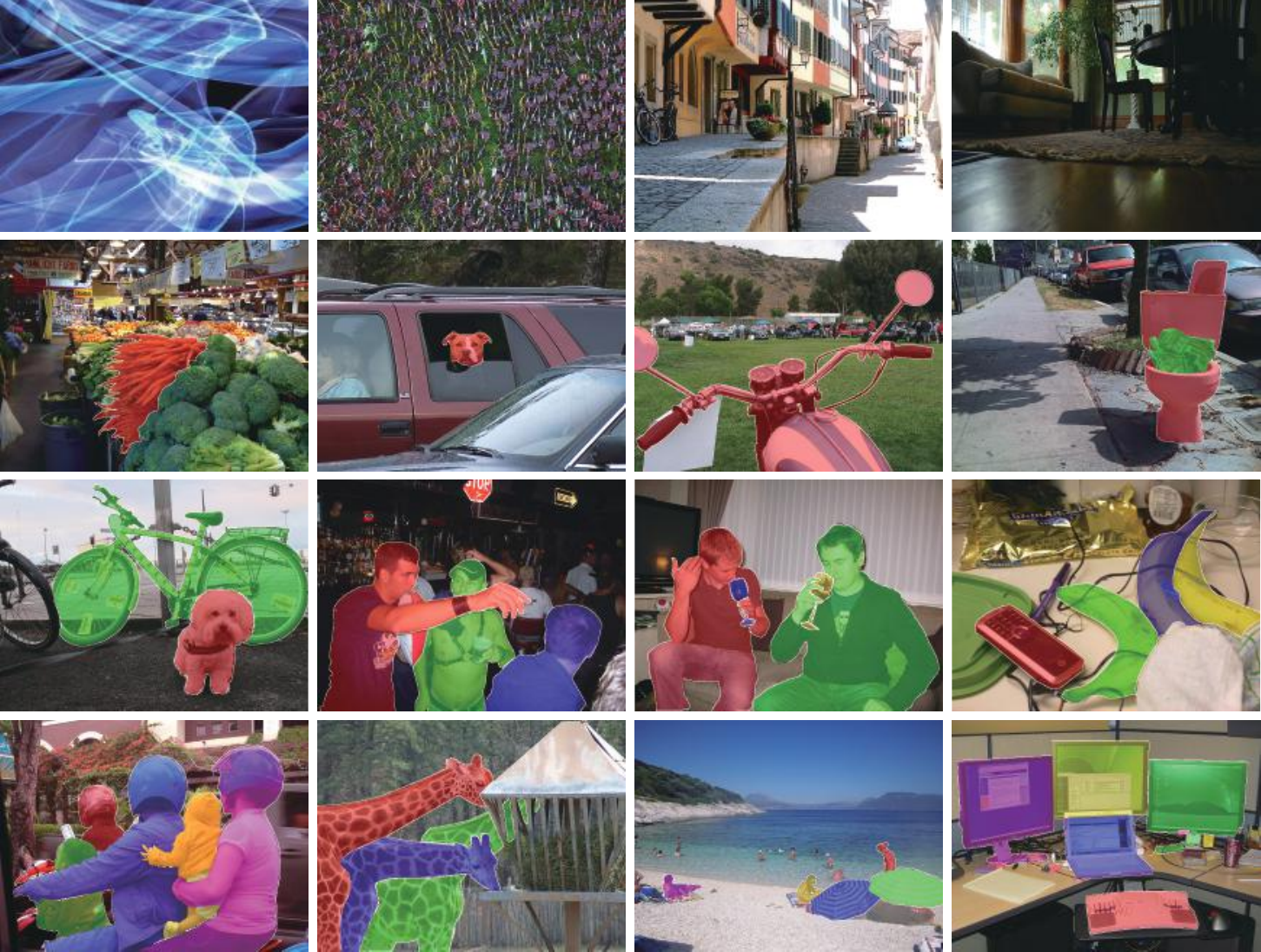}
        \AddText{8}{65}{\scriptsize {non-salient}}
        \AddText{33}{65}{\scriptsize{non-salient}}
        \AddText{58}{65}{\scriptsize {non-salient}}
        \AddText{83}{65}{\scriptsize {non-salient}}

        \AddText{8}{45}{\scriptsize {carrot}}\AddAttr{1}{53}{\tiny {HO,OC,SC,SO}}
        \AddText{33}{45}{\scriptsize {dog}}\AddAttr{26}{53}{\tiny {HO,OC,SC,SO}}
        \AddText{58}{45}{\scriptsize {motorcycle}}\AddAttr{51}{53}{\tiny {HO,SC}}
        \AddText{90}{40}{\scriptsize {toilet}}  \AddAttr{76}{53}{\tiny {HO,OC,AC}}
        \AddText{88}{48}{\scriptsize{book}}

        \AddText{13}{30}{\scriptsize {bicycle}} \AddAttr{1}{34}{\tiny {OC,SC}}
        \AddText{15}{23}{\scriptsize {dog}}
        \AddText{32}{23}{\scriptsize {person}} \AddAttr{26}{34}{\tiny {HO,OC,OV,SC}}
        \AddText{60}{26}{\scriptsize {wine glass}}
        \AddText{58}{23}{\scriptsize {person}} \AddAttr{51}{34}{\tiny {OV}}
        \AddText{79}{20}{\scriptsize {cell phone}}
        \AddText{88}{26}{\scriptsize {banana}} \AddAttr{76}{34}{\tiny {HO,OC,SC}}

        \AddText{5}{8}{\scriptsize {person}}   \AddAttr{1}{15}{\tiny {BO,OC}}
        \AddText{30}{8}{\scriptsize {giraffe}} \AddAttr{26}{15}{\tiny {HO,OC,OV}}
        \AddText{54}{3}{\scriptsize {person}}  \AddAttr{51}{15}{\tiny {HO,OC,AC}}
        \AddText{62}{1}{\scriptsize {umbrella}}
        \AddText{83}{12}{\scriptsize {computers}}\AddAttr{76}{15}{\tiny {HO}}
        \AddText{89}{7}{\scriptsize {laptop}}
        \AddText{86}{3}{\scriptsize {keyboard}}
    \end{overpic}
	\caption{Sample images from our new dataset including
    \emph{non-salient} object images (first row) and salient object images (rows 2 to 4).
    For salient object images, instance-level ground truth map (different color),
    object attributes (Attr) and category labels are provided.
    Please refer to the supplemental material for more illustrations of our dataset.}
    \label{fig:DatasetExample}
\end{figure}

\section{Introduction}
This paper considers the task of salient object detection (SOD).
Visual saliency mimics the ability of the human visual system
to select a certain subset of the visual scene. SOD aims to detect the most
attention-grabbing objects in a scene and then extract pixel-accurate
silhouettes of the objects. The merit of SOD lies in it applications in
many other computer vision tasks including: visual tracking~\cite{borji2013state},
image retrieval~\cite{he2012mobile,Fan2018Enhanced}, 
computer graphics \cite{ChengSurveyVM2017},
content aware image resizing \cite{ResizingZhangC09}, 
and weakly supervised semantic segmentation 
\cite{wei2016stc,AdversErasingCVPR2017,HouDMWCT17}.

Our work is motivated by two observations. First, existing SOD
datasets~\cite{LiuSZTS07Learn,alpert2007image,2001iccvSOD,ChengPAMI15,borji2012salient,
yan2013hierarchical,YangZLRY13Manifold,li2014secrets,li2015visual,ChengGroupSaliency}
are flawed either in the data collection procedure or quality of the data.
Specifically, most datasets assume that an image contains at least one
salient object, and thus discard images that do not contain salient objects. We
call this \emph{data selection bias}.
Moreover, existing datasets mostly contain images with a single object or several objects
(often a person) in low clutter. These datasets do not adequately reflect the complexity
of images in the real world where scenes usually contain multiple objects amidst lots of clutter.
As a result, all top performing models trained on the existing datasets have nearly saturated the
performance (\eg, $>$ 0.9 $F\textrm{-}measure$ over most current datasets) but unsatisfactory
performance on realistic scenes (\eg, $<$ 0.45 $F\textrm{-}measure$ in \tabref{tab:Attribute}).
Because current models may be biased towards ideal conditions, their effectiveness may be impaired
once they are applied to real world scenes.
To solve this problem, it is important to introduce a dataset that reaches closer to realistic conditions.

Second, only the overall performance of the models can be analyzed over existing datasets.
None of the datasets contains various attributes that reflect challenges in real-world scenes.
Having attributes helps 1) gain a deeper insight into the SOD problem, 2) investigate
the pros and cons of the SOD models, and 3) objectively assess the model performances over
different perspectives, which might be diverse for different applications.

Considering the above two issues, we make two contributions. Our main contribution is
the collection of a new high quality SOD dataset, named the \textbf{SOC},
Salient Objects in Clutter.
To date, SOC is the largest instance-level SOD dataset and contains
6,000 images from more than 80 common categories. It differs from existing datasets
in three aspects: 1) salient objects have category annotation which can be used for new research
such as weakly supervised SOD tasks, 2) the inclusion of non-salient images which
make this dataset closer to the real-world scenes and more challenging than the existing ones,
and 3) salient objects have attributes reflecting specific situations faced in the real-wold such
as \emph{motion blur, occlusion and cluttered background}.
As a consequence, our SOC dataset narrows the \emph{\textbf{gap}} between existing datasets and the
real-world scenes and provides a more realistic benchmark (see~\figref{fig:DatasetExample}).

In addition, we provide a comprehensive evaluation of several state-of-the-art convolutional
neural networks (CNNs) based models~\cite{wang2015deep,zhao2015saliency,li2015visual,li2016deep,lee2016deep,liu2016dhsnet,
wang2016saliency,chen2016disc,zhang2017deep,HouPami18Dss,Luo2017CVPR,zhang2017amulet,zhang2017learning}.
To evaluate the models, we introduce three metrics that measure the region similarity of the detection,
the pixel-wise accuracy of the segmentation, and the structure similarity of the result.
Furthermore, we give an attribute-based performance evaluation. These attributes allow a deeper
understanding of the models and point out promising directions for further research.

We believe that our dataset and benchmark can be very influential for future SOD research in particular
for application-oriented model development. The entire dataset and analyzing tools will be released freely to the public.

\section{Related Works}
In this section, we briefly discuss existing datasets designed for
SOD tasks, especially in the aspects including annotation type, the number
of salient objects per image, number of images, and image quality.
We also review the CNNs based SOD models.

\begin{figure}[thp!]
	\centering
	\begin{overpic}[width=.75\columnwidth]{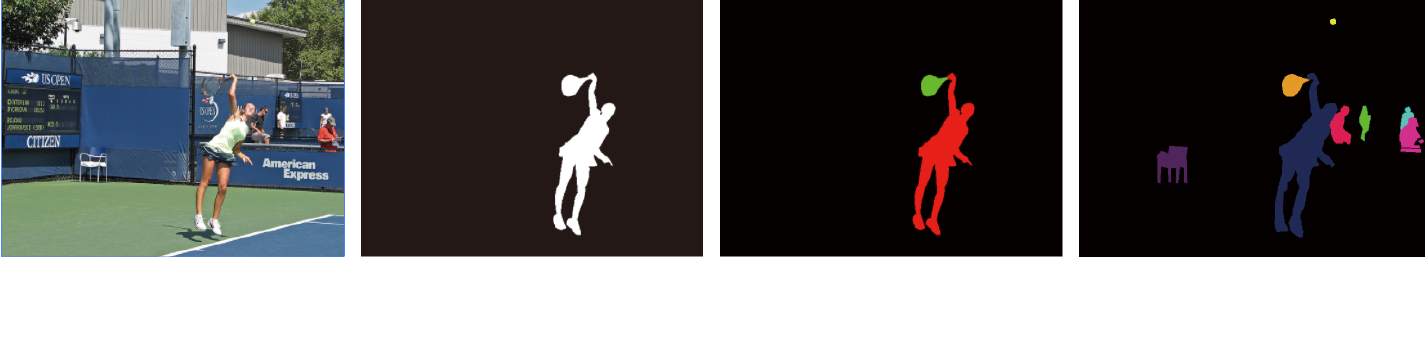}
    \put(3,2){\footnotesize {(a) Image}}
    \put(29,2){\footnotesize {(b) Pixel}}
    \put(51,2){\footnotesize {(c) Instance}}
    \put(77,2){\footnotesize {(d) Segment}}
    \end{overpic}
    \caption{Previous SOD datasets only annotate the image
    by drawing (b) pixel-accurate silhouettes of salient objects.
    Different from (d) MS COCO object segmentation dataset~\cite{lin2014microsoft}
    (Objects are not necessarily being \emph{salient}),
    our work focuses on (c) segmenting \emph{salient object instances}.
    }\label{fig:AnnotationDifference}
\end{figure}
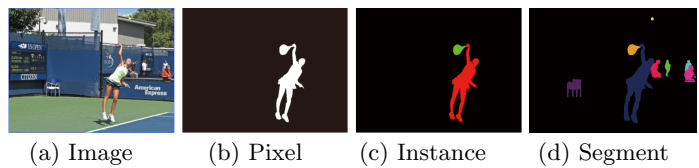

\subsection{Datasets}
Early datasets are either limited in the number of images or in their coarse annotation of salient objects.
For example, the salient objects in datasets \textbf{MSRA-A}~\cite{LiuSZTS07Learn} and \textbf{MSRA-B}
~\cite{LiuSZTS07Learn} are roughly annotated in the form of bounding boxes.
\textbf{ASD}~\cite{achanta2009frequency} and \textbf{MSRA10K}~\cite{ChengPAMI15} mostly contain
only one salient object in each image,
while the \textbf{SED2}~\cite{alpert2007image} dataset contains two objects in a single image but
contains only 100 images.
To improve the quality of datasets, researchers in recent years started to collect datasets with
multiple objects in relatively complex and cluttered backgrounds. These datasets include
\textbf{DUT-OMRON}~\cite{YangZLRY13Manifold}, \textbf{ECSSD}~\cite{yan2013hierarchical},
\textbf{Judd-A}~\cite{borji2012salient}, and \textbf{PASCAL-S}~\cite{li2014secrets}.
These datasets have been improved in terms of annotation quality and the number of images,
compared to their predecessors.
Datasets \textbf{HKU-IS}~\cite{li2015visual}, \textbf{XPIE}~\cite{xia2017and}, and \textbf{DUTS}
~\cite{wang2017learning} resolved the shortcomings by collecting large amounts of pixel-wise labeled
images (~\figref{fig:AnnotationDifference} (b)) with more than one salient object in images.
However, they ignored the non-salient objects and did not offer instance-level
(\figref{fig:AnnotationDifference} (c)) salient objects annotation.
Beyond these, researchers of~\cite{jiang2017joint} collected about 6k \textit{simple background
images} (most of them are pure texture images) to account for the non-salient scenes.
This dataset is not sufficient to reflect real scenes as the real-world scenes are more
complicated. The \textbf{ILSO}~\cite{li2017instance} dataset contains instance-level salient
objects annotation but has boundaries roughly labeled as shown in \figref{fig:HighQualityAnnotation} (a).

To sum up, as discussed above, existing datasets mostly focus on images with clear salient objects
in simple backgrounds. Taking into account the aforementioned limitations of existing datasets, a
more realistic dataset which contains realistic scenes with non-salient objects,
textures ``in the wild'', and salient objects with attributes, is needed for future investigations
in this field. Such a dataset can offer deep insights into weaknesses and strengths of SOD models.

\subsection{Models}
We divide the \sArt~deep models for SOD based on the number of tasks.

\textbf{Single-task} models have the single goal of detecting the salient objects in images.
In \textbf{LEGS}~\cite{wang2015deep}, local information and global contrast were
separately captured by two different deep CNNs, and were then fused to generate a saliency map.
In \cite{zhao2015saliency}, Zhao \etal~presented a multi-context deep learning 
framework (\textbf{MC}) for SOD.
Li \etal~\cite{li2015visual} (\textbf{MDF}) proposed to use
multi-scale features extracted from a deep CNNs to derive a saliency map.
Li \etal~\cite{li2016deep} presented a deep contrast network
(\textbf{DCL}), which not only considered the pixel-wise information
but also fused the segment-level guidance into the network.
Lee \etal~\cite{lee2016deep} (\textbf{ELD}) considered both high-level
features extracted from CNNs and hand-crafted features.
Liu \etal~\cite{liu2016dhsnet} (\textbf{DHS}) designed a two-stage network, in which a coarse
downscaled prediction map was produced. It is then followed by another network to refine the details
and upsample the prediction map hierarchically and progressively.
Long \etal~\cite{long2015fully} proposed a fully convolutional network (\textbf{FCN})
to make dense pixel prediction problem feasible for end-to-end training.
\textbf{RFCN}~\cite{wang2016saliency} used a recurrent FCN to incorporate
the coarse predictions as saliency priors and refined the
generated predictions in a stage-wise manner.
The \textbf{DISC}~\cite{chen2016disc} framework was proposed for fine-grained image saliency computing.
Two stacked CNNs were utilized to obtain coarse-level and fine-grained saliency maps, respectively.
\textbf{IMC}~\cite{zhang2017deep} integrated saliency cues at different levels through FCN.
It could efficiently exploit both learned semantic cues and higher-order region statistics for edge-accurate SOD.
Recently, a deep architecture \cite{HouPami18Dss} with short connections (\textbf{DSS}) was proposed.
Hou \etal~added connections from high-level features to low-level features based on the
HED~\cite{xie2015holistically} architecture, achieving good performance.
\textbf{NLDF}~\cite{Luo2017CVPR} integrated local and global features and added a boundary
loss term into standard cross entropy loss to train an end-to-end network.
\textbf{AMU}~\cite{zhang2017amulet} was a generic aggregating multi-level convolutional feature framework.
It integrated coarse semantics and fine detailed feature maps into multiple resolutions.
Then it adaptively learned to combine these feature maps at each resolution
and predicted saliency maps with the combined features.
\textbf{UCF}~\cite{zhang2017learning} was proposed to improve the robustness and accuracy of saliency detection.
They introduced a reformulated dropout after specific convolutional layers to construct an uncertain ensemble
of internal feature units.
Also, they proposed reformulated dropout after an effective hybrid up-sampling method to reduce the
checkerboard artifacts of deconvolution operators in the decoder network.

\begin{table*}[t!]
  \centering
  \scriptsize
  \renewcommand{\arraystretch}{0.5}
  \renewcommand{\tabcolsep}{0.65mm}

  \caption{CNNs based SOD models.
  We divided these models into single-task (S-T) and multi-task (M-T).
  \textbf{\emph{Training Set:}}
  MB is the MSRA-B dataset~\cite{LiuSZTS07Learn}.
  MK is the MSRA-10K~\cite{ChengPAMI15} dataset.
  ImageNet dataset refers to~\cite{russakovsky2015imagenet}.
  D is the DUT-OMRON~\cite{YangZLRY13Manifold} dataset.
  H is the HKU-IS~\cite{li2015visual} dataset.
  P is the PASCAL-S~\cite{li2014secrets} dataset.
  P2010 is the PASCAL VOC 2010 semantic segmentation dataset~\cite{pascal-voc-2010}.
  \textbf{\emph{Base Model:}}
  VGGNet, ResNet-101, AlexNet, GoogleNet are base models.
  \textbf{\emph{FCN:}} whether model uses the fully convolutional network.
  \textbf{\emph{Sp:}} whether model uses superpixels.
  \textbf{\emph{Proposal:}} whether model uses the object proposal.
  \textbf{\emph{Edge:}} whether model uses the edge or contour information
  }\label{tab:CNNModelSummary}

  \begin{tabular}{c|c|l|c|l|c|c|c|cccc}
  \toprule
   &No  & Model & Year & Pub & \#Training & \emph{\textbf{Training Set}} & \emph{\textbf{Base Model}} & \emph{\textbf{FCN}} & \emph{\textbf{Sp}} & \emph{\textbf{Proposal}} & \emph{\textbf{Edge}} \\
  \midrule
  \multirow{21}{*}{\begin{sideways}S-T\end{sideways}}
  &1& \textbf{LEGS}~\cite{wang2015deep} & 2015&CVPR & 3,340 & MB + P & --- &\tickNo&\tickNo&\cmark&\tickNo \\
  &2& \textbf{MC}~\cite{zhao2015saliency} & 2015&CVPR& 8,000 & MK & GoogLeNet&\tickNo&\cmark&\tickNo&\tickNo\\
  &3& \textbf{MDF}~\cite{li2015visual} & 2015& CVPR& 2,500 & MB & --- &\tickNo&\cmark&\tickNo&\cmark\\
  &4& \textbf{DCL}~\cite{li2016deep} & 2016&CVPR & 2,500 & MB & VGGNet  & \cmark&\cmark&\tickNo&\tickNo\\
  &5& \textbf{ELD}~\cite{lee2016deep}& 2016& CVPR& 9,000& MK &VGGNet&\tickNo&\cmark&\tickNo&\tickNo\\
  &6& \textbf{DHS}~\cite{liu2016dhsnet} & 2016&CVPR & 9,500 & MK+D & VGGNet&\tickNo&\tickNo&\tickNo&\tickNo\\
  &7& \textbf{RFCN}~\cite{wang2016saliency}& 2016 & ECCV & 10,103 & P2010 & --- &\cmark&\cmark&\tickNo&\cmark\\
  &8& \textbf{DISC}~\cite{chen2016disc} & 2016 & TNNLS & 9,000 & MK&---&\tickNo&\cmark&\tickNo&\tickNo\\
  &9& \textbf{IMC}~\cite{zhang2017deep} & 2017&WACV & 6,000 & MK &ResNet-101&\cmark &\cmark &\tickNo&\tickNo\\
  &10& \textbf{DSS}~\cite{HouPami18Dss} & 2017&CVPR & 2,500 & MB & VGGNet&\cmark&\tickNo&\tickNo&\cmark\\
  &11& \textbf{NLDF}~\cite{Luo2017CVPR} & 2017&CVPR & 2,500 & MB &VGGNet&\cmark&\tickNo&\tickNo&\tickNo\\
  &12& \textbf{AMU}~\cite{zhang2017amulet} & 2017&ICCV & 10,000 & MK &VGGNet&\cmark&\tickNo&\tickNo&\cmark\\
  &13& \textbf{UCF}~\cite{zhang2017learning} & 2017&ICCV & 10,000 & MK &---&\cmark & \tickNo &\tickNo&\tickNo\\
  \midrule

  \multirow{3}{*}{\begin{sideways}M-T \end{sideways}}
  &1& \textbf{DS}~\cite{li2016deepsaliency} & 2016&TIP & 10,000 &MK&VGGNet&\cmark&\cmark &\tickNo&\tickNo\\
  &2& \textbf{WSS}~\cite{wang2017learning} & 2017&CVPR & 456K & ImageNet &VGGNet&\cmark&\cmark&\tickNo&\tickNo\\
  &3& \textbf{MSR}~\cite{li2017instance} & 2017&CVPR & 5,000 & MB + H &VGGNet&\cmark&\tickNo&\cmark&\cmark\\
  \bottomrule
  \end{tabular}
\end{table*}

\textbf{Multi-task} models at present include three methods, \textbf{DS}, \textbf{WSS}, and
\textbf{MSR}. The \textbf{DS}~\cite{li2016deepsaliency} model set up a multi-task learning scheme
for exploring the intrinsic correlations between saliency detection and semantic image
segmentation, which shared the information in FCN layers to generate effective features for object
perception. Recently, Wang \etal~\cite{wang2017learning} proposed a model named \textbf{WSS} which
developed a weakly supervised learning method using image-level tags for saliency detection.
First, they jointly trained Foreground Inference Net (FIN) and FCN for image categorization.
Then, they used FIN fine-tuned with iterative CRF to enforce spatial label consistency to predict
the saliency map. \textbf{MSR}~\cite{li2017instance} was designed for both salient region detection
and salient object contour detection, integrated with multi-scale combinatorial grouping and a
MAP-based~\cite{zhang2016unconstrained} subset optimization framework. Using three refined VGG
network streams with shared parameters and a learned attentional model for fusing results at
different scales, the authors were able to achieve good results.

We benchmark a large set of the state-of-the-art CNNs based models (see \tabref{tab:CNNModelSummary})
on our proposed dataset, highlighting the current issues and pointing out future research directions.

\section{The Proposed Dataset}\label{sec:our_dataset}
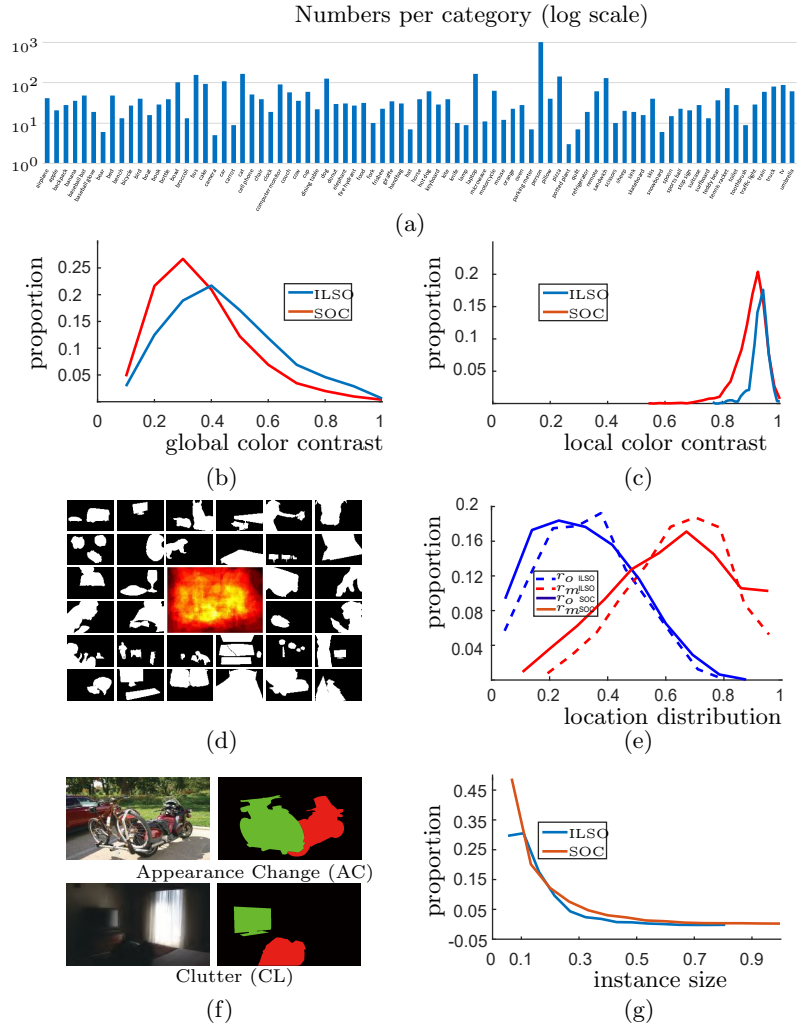
\begin{figure}[t!]
	\centering
	\begin{overpic}[width=.86\columnwidth]{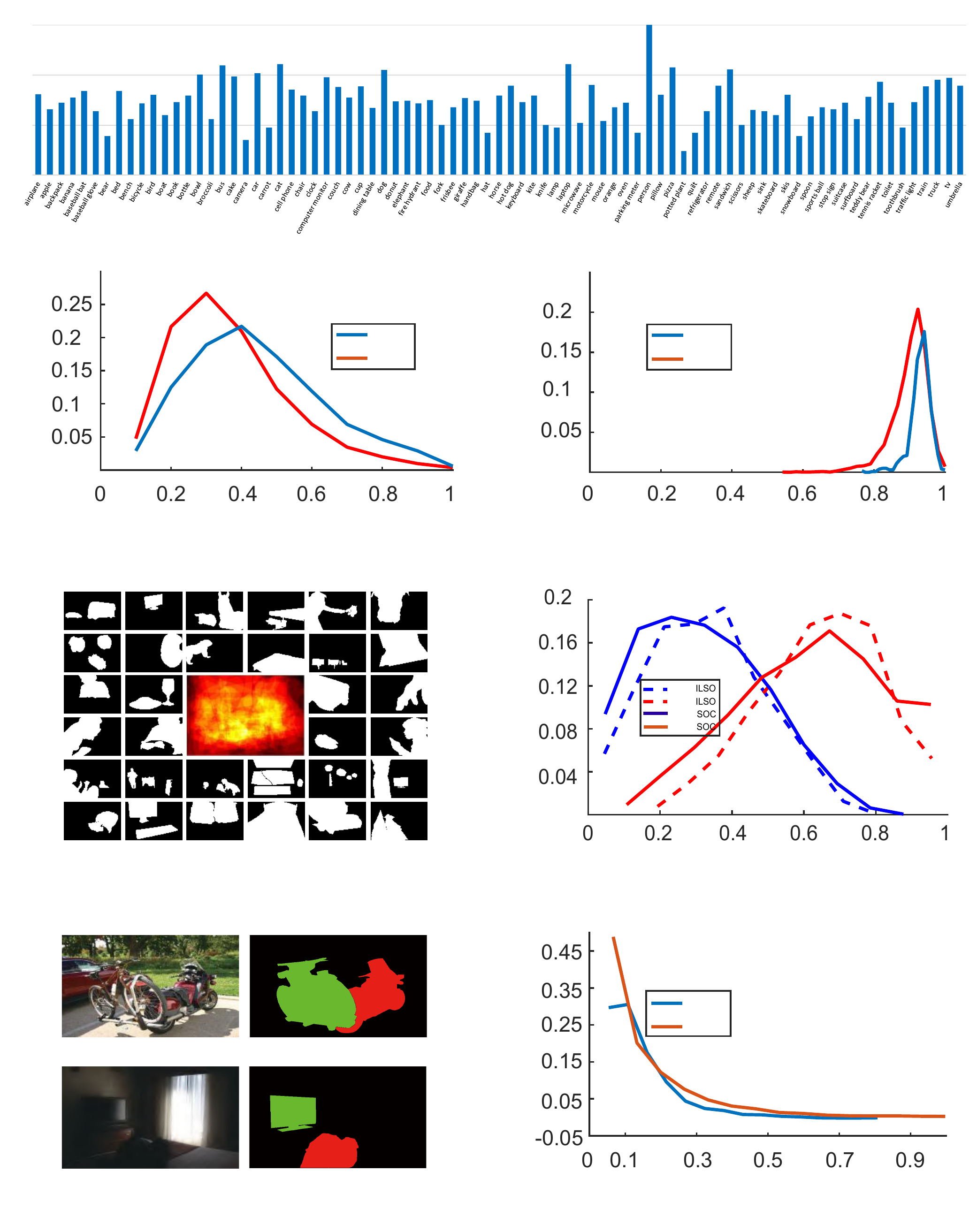}
    \put (28,100){\footnotesize Numbers per category (log scale)}
    \put (-0.5,85){\tiny {$10^0$} }
    \put (-0.5,89){\tiny {$10^1$} }
    \put (-0.5,93){\tiny {$10^2$} }
    \put (-0.5,97){\tiny {$10^3$} }
    \put(38,79){\small (a)}

    \put(19,53.5){\small (b)}
    \put(29.9,72.2){\tiny {ILSO}}
    \put(29.9,70.3){\tiny  {SOC}}
    \put(1,65){\rotatebox{90}{proportion}}
    \put(15,57){\footnotesize{global color contrast}}

    \put(61,53.5){\small (c)}
    \put(55.5,72.2){\tiny {ILSO}}
    \put(55.5,70.3){\tiny  {SOC}}
    \put(41,65){\rotatebox{90}{proportion}}
    \put(55,57){\footnotesize{local color contrast}}

    \put(19,27){\small (d)}

    \put(61,27){\small (e)}
    \put(55,29.4){\footnotesize{location distribution}}
    \put(41,39){\rotatebox{90}{proportion}}
    \put(54.2,44.0){\tiny {$r_{o}$}}
    \put(54.2,42.8){\tiny {$r_{m}$}}
    \put(54.2,41.8){\tiny {$r_{o}$}}
    \put(54.2,40.8){\tiny {$r_{m}$}}

    \put(19,0){\small (f)}
    \put(12,14){\scriptsize {Appearance Change (AC)}}
    \put(16,3.5){\scriptsize {Clutter (CL)}}

    \put(61,0){\small (g)}
    \put(55.5,18){\tiny {ILSO}}
    \put(55.5,16.2){\tiny  {SOC}}
    \put(58,3){\footnotesize{instance size}}
    \put(41,10){\rotatebox{90}{proportion}}

    \end{overpic}
    \caption{(a) Number of annotated instances per category in our SOC dataset.
    (b, c) The statistics of global color contrast and local color contrast, respectively.
    (d) A set of saliency maps from our dataset and their overlay map. (e) Location distribution of the
    salient objects in SOC. (f) Attribute visual examples.
    (g) The distribution of instance sizes for the SOC and ILSO~\cite{li2017instance}.
    }\label{fig:NewStatistical}
\end{figure}

\begin{figure}[t!]
\begin{center}
    \includegraphics[width=.6\columnwidth]{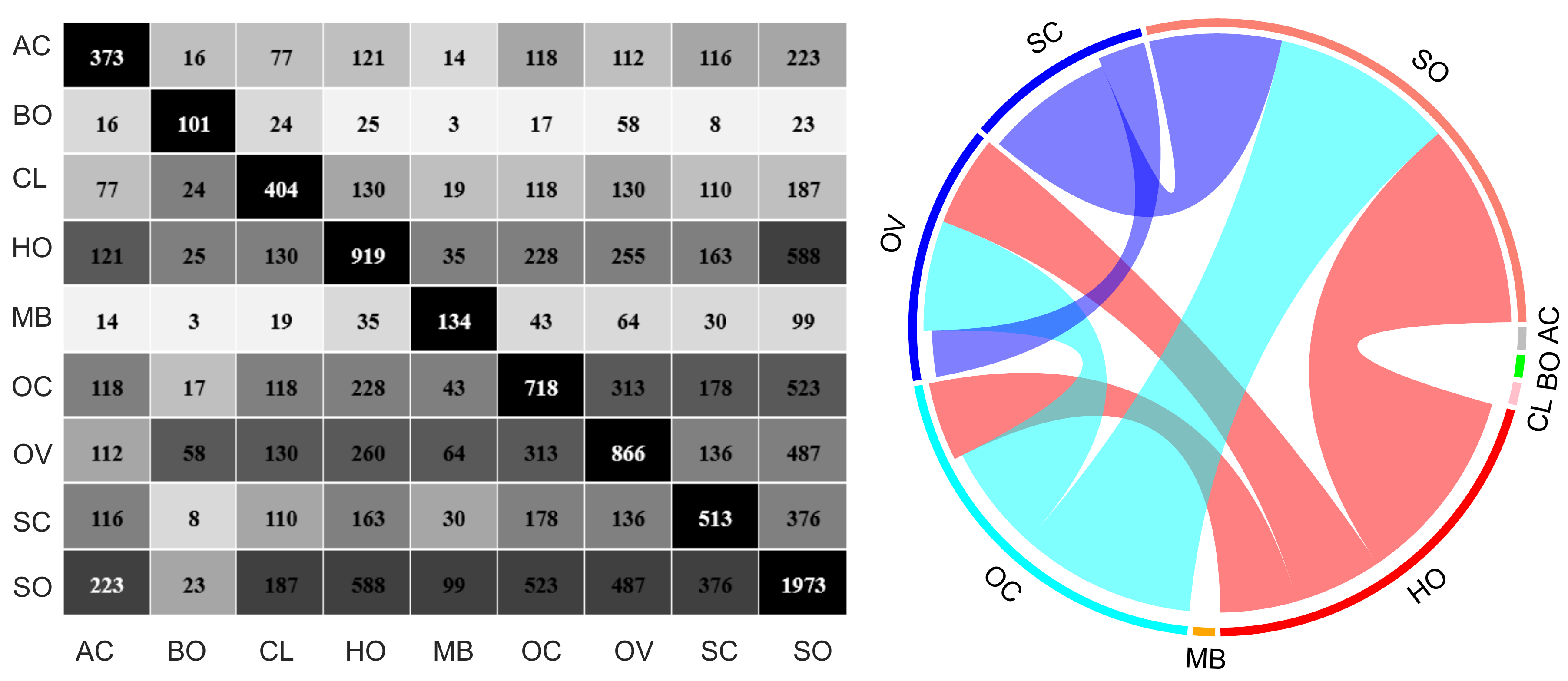}
     \caption{Left: Attributes distribution over the salient object images in our SOC dataset.
     Each number in the grids indicates the image number of occurrences.
     Right: The \emph{dominant dependencies} among attributes base on the frequency of occurrences.
     Larger width of a link indicates higher probability of an attribute to other ones.}\label{fig:Attributes_distribution}
\end{center}
\end{figure}

In this section, we present our new challenging SOC dataset
designed to reflect the real-world scenes in detail.
Sample images from SOC are shown in \figref{fig:DatasetExample}.
Moreover, statistics regarding the categories and the attributes of SOC are shown in
Fig.~\ref{fig:NewStatistical} (a) and \figref{fig:Attributes_distribution}, respectively.
Based on the strengths and weaknesses of the existing datasets,
we identify seven crucial aspects that a comprehensive and balanced dataset should fulfill.

\begin{figure}[thp!]
  \centering
  \begin{overpic}[width=.7\columnwidth]{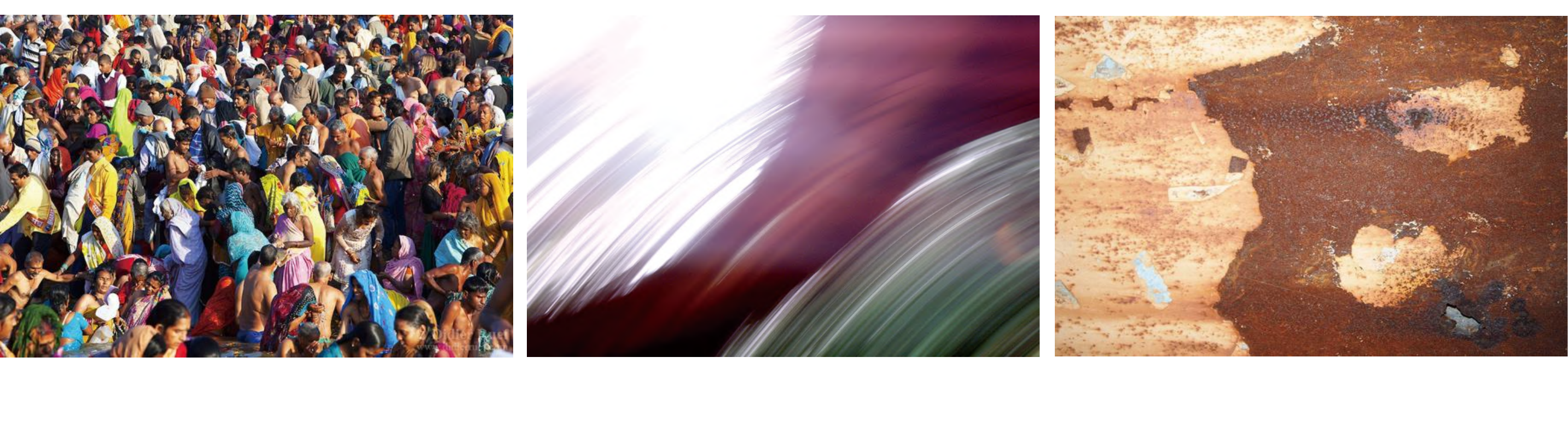}
  \put(14,0){\small {(a)}}
  \put(46,0){\footnotesize {(b)}}
  \put(83,0){\footnotesize {(c)}}
  \end{overpic}
  \caption{Some examples of non-salient objects.}\label{fig:nonsal}
\end{figure}

\textbf{1) Presence of Non-Salient Objects.}
Almost all of the existing SOD datasets
make the assumption that an image contains at least one salient object and
discard the images that do not contain salient objects.
However, this assumption is an ideal setting which leads to \emph{data selection bias}.
In a realistic setting, images do not always contain salient objects.
For example, some amorphous background images such as sky, grass and texture
contain no salient objects at all~\cite{Coco2018stuff}.
The non-salient objects or background ``stuff'' may occupy the entire scene,
and hence heavily constrain possible locations for a salient object.
Xia \etal~\cite{xia2017and} proposed a \sArt~SOD model
by judging what is or what is not a salient object, indicating that the non-salient object is crucial
for reasoning about the salient object.
This suggests that the non-salient objects deserve equal attention as the salient objects in SOD.
Incorporating a number of images containing non-salient objects makes
the dataset closer to real-world scenes, while becoming more challenging.
Thus, we define the ``\emph{non-salient objects}''
as images without salient objects or images with ``stuff'' categories.
As suggested in~\cite{Coco2018stuff,xia2017and},
the ``stuff'' categories including (a) densely distributed similar objects,
(b) fuzzy shape, and (c) region without semantics,
which are illustrated in \figref{fig:nonsal} (a)-(c), respectively.

Based on the characteristics of non-salient objects, we collected 783 texture images from the DTD~\cite{lazebnik2005sparse} dataset.
To enrich the diversity, 2217 images including aurora, sky,
crowds, store and many other kinds of realistic scenes were gathered from the Internet and other datasets~\cite{jiang2013salient,lin2014microsoft,li2014secrets,2001iccvSOD}.
We believe that incorporating enough non-salient objects would open up a promising direction for future works.

\textbf{2) Number and Category of Images.}
A considerably large amount of images is essential to capture the diversity
and abundance of real-world scenes.
Moreover, with large amounts of data, SOD models can avoid over-fitting and enhance generalization.
To this end, we gathered 6,000 images from more than 80 categories,
containing 3,000 images with salient objects and 3,000 images without salient
objects.
We divide our dataset into training set,
validation set and test set in the ratio of 6:2:2.
To ensure fairness, the test set is not published,
but with the \emph{on-line testing} provided on our website\footnote{
\url{http://dpfan.net/SOCBenchmark/}}.
\figref{fig:NewStatistical} (a) shows the number of salient objects for each category.
It shows that the ``person'' category accounts for a large proportion,
which is reasonable as people usually
appear in daily scenes along with other objects.

\textbf{3) Global/Local Color Contrast of Salient Objects.}
As described in~\cite{li2014secrets},
the term ``salient'' is related to the global/local contrast of the foreground
and background. It is essential to check whether the salient objects
are easy to detect. For each object, we compute RGB color histograms for
foreground and background separately. Then, $\chi^2$ distance is utilized to
measure the distance between the two histograms.
The global and local color contrast distribution
are shown in \figref{fig:NewStatistical} (b) and (c), respectively.
In comparison to ILSO, our SOC has more proportion of objects with low global
color contrast and local color contrast.

\textbf{4) Locations of Salient Objects.}
\emph{Center bias} has been identified as one of the most significant
biases of saliency detection datasets
\cite{borji2015salient,li2014secrets,Judd_2012}.
\figref{fig:NewStatistical} (d) illustrates a set of images and their overlay map.
As can be seen, although salient objects are located in different positions,
the overlay map still shows that somehow this set of images is center biased.
Previous benchmarks often adopt this incorrect way to analyze the location
distribution of salient objects.
To avoid this misleading phenomenon, we plot the statistics of two quantities $r_o$ and $r_m$
in \figref{fig:NewStatistical} (e),
where $r_o$ and $r_m$ denote how far an object center
and the farthest (margin) point in an object are
from the image center, respectively.
Both $r_o$ and $r_m$ are divided by half image diagonal length
for normalization so that $r_o, r_m \in [0, 1]$.
From these statistics,
we can observe that salient objects in our dataset do not suffer
from center bias.

\textbf{5) Size of Salient Objects.}
The size of an instance-level salient object
is defined as the proportion of pixels in the image~\cite{li2014secrets}.
As shown in \figref{fig:NewStatistical} (g), the size of salient objects
in our SOC varies in a broader range, compared with
the only existing instance-level ILSO~\cite{li2017instance} dataset.
Also, medium-sized objects in SOC have a higher proportion.

\begin{figure}[thp!]
  \centering
  \begin{overpic}[width=.6\columnwidth]{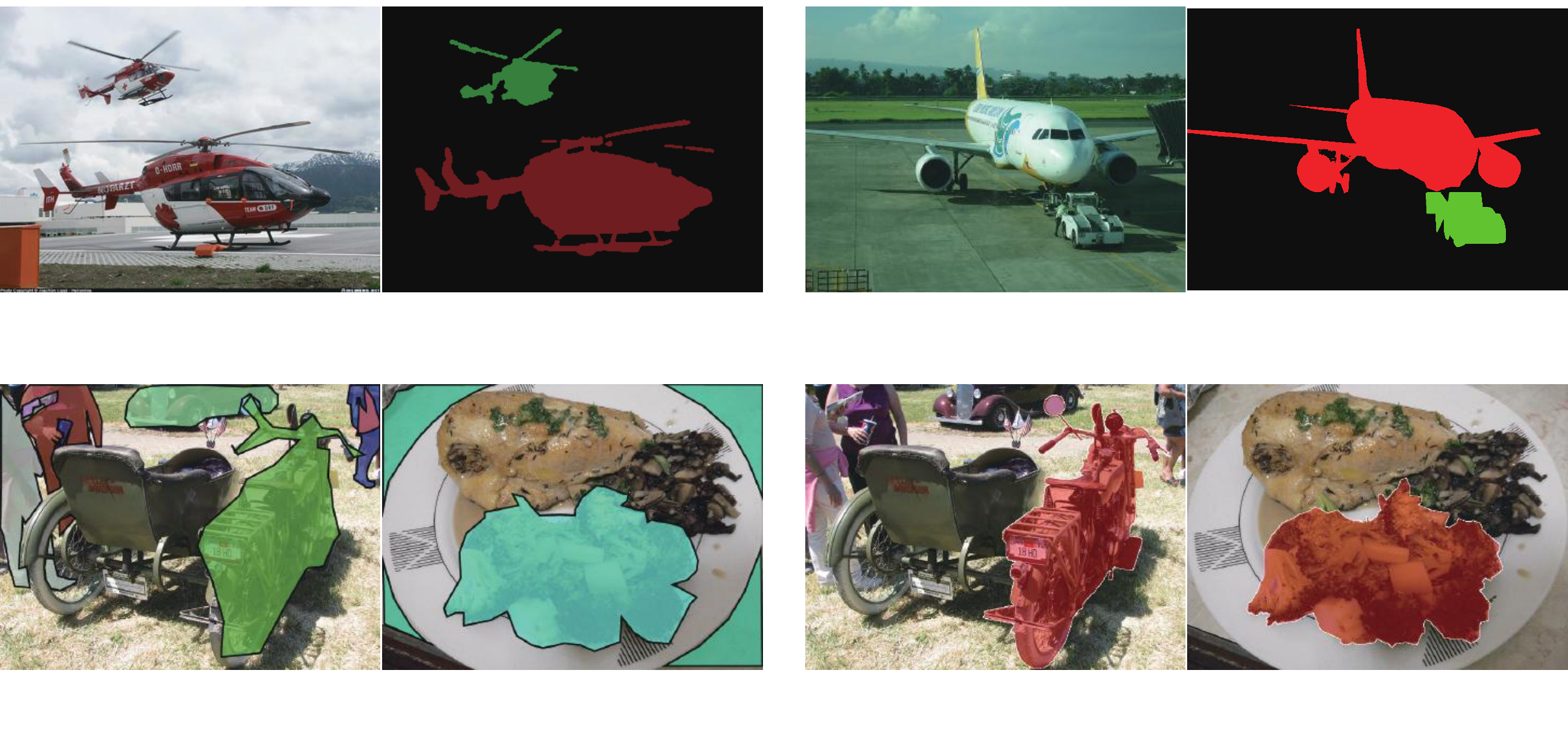}
  \put(16,26){\footnotesize {(a) ILSO} }
  \put(68,26){\footnotesize {(b) SOC} }
  \put(13,2){\footnotesize {(c) MSCOCO} }
  \put(68,2){\footnotesize {(d) SOC} }
  \end{overpic}
  \caption{Compared with the recent new (a) instance-level ILSO dataset~\cite{li2017instance}
    which is labeled with discontinue coarse boundaries, (c) MSCOCO dataset~\cite{lin2014microsoft}
    which is labeled with polygons, our (b, d) SOC dataset is labeled with  smooth fine boundaries.
  }\label{fig:HighQualityAnnotation}
\end{figure}
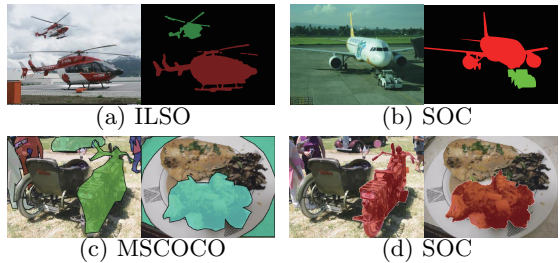

\textbf{6) High-Quality Salient Object Labeling.}
As also noticed in \cite{HouPami18Dss}, training on
the ECSSD dataset (1,000) allows to achieve better results than
other datasets (\eg, MSRA10K, with 10,000 images).
Besides the scale, dataset quality is also an important factor.
To obtain a large amount of high quality images,
we randomly select images from the MSCOCO dataset~\cite{lin2014microsoft},
which is a large-scale real-world dataset
whose objects are labeled with polygons (\ie, coarse labeling).
High-quality labels also play a
critical role in improving the accuracy of SOD models \cite{achanta2009frequency}.
Toward this end, we relabel the dataset with pixel-wise annotations.
Similar to famous SOD task oriented benchmark datasets~\cite{achanta2009frequency,alpert2007image,ChengPAMI15,
jiang2017joint,li2017instance,li2015visual,LiuSZTS07Learn,2001iccvSOD,wang2017learning,
xia2017and,yan2013hierarchical}, we did not use the eye tracker device.
We have taken a number of steps to provide
the high-quality of the annotations.
These steps include two stages: \textbf{In the bounding boxes (bboxes) stage}, (i)
we ask 5 viewers to annotate objects with bboxes that they think are
salient in each image.
(ii) keep the images which majority ($\geq 3$) viewers
annotated the same (the IOU of the bbox $>0.8$) object.
After the first stage, we have
3,000 salient object images annotated with bboxes.
\textbf{In the second stage}, we further manually label the accurate silhouettes
of the salient objects according
to the bboxes. Note that we have 10 volunteers involved in the whole steps for
cross-check the quality of annotations.
In the end, we keep 3,000 images with high-quality, instance-level labeled
salient objects. As shown in \figref{fig:HighQualityAnnotation} (b,d),
the boundaries of our object labels are precise, sharp and smooth. During the
annotation process, we also add some new categories
(\eg, \emph{computer monitor, hat, pillow}) that are not labeled in
the MSCOCO dataset~\cite{lin2014microsoft}.

\begin{table}[t!]
\scriptsize
\caption{The list of salient object image attributes and the corresponding description.
By observing the characteristics of the existing datasets, we summarize these attributes.
Some visual examples can be found in \figref{fig:DatasetExample} and \figref{fig:NewStatistical} (f).
For more examples, please refer to the supplementary materials}\label{tab:Attr}
\begin{tabular*}{\linewidth}{ll}
  \toprule
  Attr & Description \\
  \midrule
  AC   & \emph{\textbf{Appearance Change.}} The obvious illumination change in the object region.\\
  BO   & \emph{\textbf{Big Object.}} The ratio between the object area and the image area is larger than 0.5.\\
  CL   & \emph{\textbf{Clutter.}} The foreground and background regions around the object have similar color.\\
       &We labeled images that their global color contrast value is larger than 0.2, local color \\
       &contrast value is smaller than 0.9 with clutter images (see \secref{sec:our_dataset}).\\
  HO   & \emph{\textbf{Heterogeneous Object.}} Objects composed of visually distinctive/dissimilar parts.\\
  MB   & \emph{\textbf{Motion Blur.}} Objects have fuzzy boundaries due to shake of the camera or motion.\\
  OC   & \emph{\textbf{Occlusion.}} Objects are partially or fully occluded.\\
  OV   & \emph{\textbf{Out-of-View.}} Part of object is clipped by image boundaries.\\
  SC   & \emph{\textbf{Shape Complexity.}} Objects have complex boundaries such as thin parts\\
       &(\eg, the foot of animal) and holes.\\
  SO   & \emph{\textbf{Small Object.}} The ratio between the object area and the image area is smaller than 0.1.\\
  \bottomrule
  \end{tabular*}
\end{table}

\textbf{7) Salient Objects with Attributes.}
Having attributes information regarding the images in a dataset helps
objectively assess the performance of models
over different types of parameters and variations.
It also allows the inspection of model failures.
To this end, we define a
set of attributes to represent specific situations
faced in the real-wold scenes such as \emph{motion blur},
\emph{occlusion} and \emph{cluttered background}
(summarized in \tabref{tab:Attr}).
Note that one image can be annotated with multiple attributes as these attributes are not exclusive.

Inspired by~\cite{perazzi2016benchmark}, we present
the distribution of attributes over the dataset as shown in \figref{fig:Attributes_distribution} Left.
Type \emph{SO} has the largest proportion due to accurate instance-level
(\eg, tennis racket in \figref{fig:AnnotationDifference}) annotation.
Type \emph{HO} accounts for a large proportion, because the real-world scenes are composed of different constituent materials.
\emph{Motion blur} is more common in video frames than still images, but it also occurs in still images
sometimes. Thus, type \emph{MB} takes a relatively small proportion in our dataset.
Since a realistic image usually contains multiple attributes, we show
the dominant dependencies among attributes based on the frequency of occurrences in the
\figref{fig:Attributes_distribution} Right.
For example, a scene containing lots of heterogeneous objects is likely to
have a large number of objects blocking each other and forming complex spatial structures.
Thus, type \emph{HO} has a strong dependency with type \emph{OC}, \emph{OV}, and \emph{SO}.

\section{Benchmarking Models}
In this section, we present the evaluation results of the sixteen
SOD models on our SOC dataset.
Nearly all representative CNNs based SOD models are evaluated.
However, since the codes of some models are not publicly available, we do not consider them here.
In addition, most models are not optimized for non-salient objects detection.
Thus, to be fair, we only use the test set of our SOC dataset to evaluate SOD models.
We describe the evaluation metrics in \secref{sec:metrics}.
Overall model performance on SOC dataset is presented in \secref{sec:metrics_statistics}
and summarized in \tabref{tab:Attribute}, while the attribute level performance
(\eg, performance of the appearance changes)
is discussed in \secref{sec:Attributes_evaluation} and summarized in \tabref{tab:Atr_Performance}.
The evaluation scripts are publicly available, and on-line evaluation test is provided on our website.

\subsection{Evaluation Metrics}\label{sec:metrics}
In a supervised evaluation framework, given a predicted map $M$
generated by a SOD model and a ground truth mask $G$,
the evaluation metrics are expected to tell which model generates the best result.
Here, we use three different evaluation metrics to evaluate SOD models on our SOC dataset.

\textbf{Pixel-wise Accuracy $\varepsilon$.} The region similarity
evaluation measure does not consider the true negative saliency assignments.
As a remedy, we also compute the normalized ([0,1]) mean absolute error (MAE)
between $M$ and $G$, defined as:

\begin{equation}\label{equ:MAE}
    \varepsilon = \frac{1}{W \times H} \sum_{x=1}^{W}\sum_{y=1}^{H} || M(x,y) - G(x,y)||,
\end{equation}

where $W$ and $H$ are the width and height of images, respectively.

\textbf{Region Similarity $F$.} To measure how well the
regions of the two maps match, we use the $F\textrm{-}measure$, defined as:
\begin{equation}\label{equ:F1_measure}
    F = \frac{(1+\beta^{2}) Precision  \times  Recall}{\beta^{2} Precision + Recall},
\end{equation}
where $\beta^{2}=0.3$ is suggested by~\cite{achanta2009frequency}
to trade-off the $recall$ and $precision$.
However, the black (all-zero matrix) ground truth is not well defined in $F\textrm{-}measure$ when
calculating $recall$ and $precision$. Under this circumstances, different foreground maps get the same
result $0$, which is apparently unreasonable.
Thus, $F\textrm{-}measure$ is not suitable for measuring the results of non-salient object detection.

However, both metrics of $\varepsilon$ and $F$ are based on pixel-wise errors and often
ignore the structural similarities. Behavioral vision studies have shown that the human visual system
is highly sensitive to structures in scenes~\cite{fan2017structure}.
In many applications, it is desired that the results of the SOD model retain the structure of objects.

\textbf{Structure Similarity $S$.}
$S\textrm{-}measure$ proposed by Fan \etal~\cite{fan2017structure} evaluates the structural similarity,
by considering both regions and objects.
Therefore, we additionally use $S\textrm{-}measure$ to evaluate the structural
similarity between $M$ and $G$.
Note that the next overall performance we evaluated and analyzed are based on the $S\textrm{-}measure$.

\begin{table*}[t!]
  \centering
  \scriptsize
  \renewcommand{\arraystretch}{1}
  \renewcommand{\tabcolsep}{0.45mm}

  \caption{The performance of SOD models under three metrics.
  $F$ stands for region similarity,
  $\varepsilon$ is the mean absolute error, and
  $S$ is the structure similarity.
  $\uparrow$ stand for the higher the number the better,
  and vice versa for $\downarrow$.
  The evaluation results are calculated according to Eqn. \eqref{equ:Avg_score} over
  our SOC dataset.
  $S_{all}, F_{all}, \varepsilon_{all}$ indicate the overall performance using the metric of $S, F,~\varepsilon$, respectively.
  \textbf{Bold} for the best.}\label{tab:Attribute}

  \begin{tabular}{rlccccccccccccccc}
  \toprule
  \multicolumn{1}{c}{} & \multicolumn{13}{c}{Single-task} & \multicolumn{3}{c}{Multi-task} \\
  \cmidrule(l){1-1}      \cmidrule(l){2-14}              \cmidrule(l){15-17}
    \rowcolor{red!10}
    \textbf{\emph{Type}} & LEGS & MC & MDF & DCL & AMU & RFCN & DHS & ELD & DISC & IMC & UCF &
    DSS & NLDF & DS & WSS &MSR\\
    & \cite{wang2015deep} & \cite{zhao2015saliency} & \cite{li2015visual} & \cite{li2016deep} & \cite{li2016deep} & \cite{wang2016saliency} & \cite{liu2016dhsnet} & \cite{lee2016deep} & \cite{chen2016disc} & \cite{zhang2017deep} & \cite{zhang2017learning} & \cite{HouPami18Dss} & \cite{Luo2017CVPR} & \cite{li2016deepsaliency} & \cite{wang2017learning} & \cite{li2017instance} \\ 
    \cmidrule(l){1-1}      \cmidrule(l){2-14}              \cmidrule(l){15-17}

    $F_{all} \uparrow$ &.276 & .291 &.307& .339 & .341 & \textbf{.435} &.360 & .317 & .288 & .352 & .333 & .341 &.352 & .347 & .327 &\textbf{.380}\\
    $S_{all} \uparrow$  & .677 & .757 & .736 & .771 & .737 &.814 & .804 & .776 & .737 & .664 & .657 &.807 & \textbf{.818} & .779 & .785 & \textbf{.819}\\
    $\varepsilon_{all} \downarrow$ &.230 & .138 & .150 & .157 & .185 & .113 & .118 & .135 & .173 & .269 & .282 &.111 & \textbf{.104} & .155 & .133 & .113\\
  \bottomrule
  \end{tabular}
\end{table*}

\subsection{Metric Statistics}\label{sec:metrics_statistics}
To obtain an overall result, we average the scores of the evaluation metrics $\eta$
($\eta\in \{F, \varepsilon, S\}$), denoted by:
\begin{equation}\label{equ:Avg_score}
    M_{\eta}(D)=\frac{1}{|D|}\sum_{I\in D}{\bar{\eta}(I_{i})},
\end{equation}
where $\bar{\eta}(I_{i})$ is the evaluation score of the image $I_i$
within the image dataset $D$.

\vspace{5pt}
\textbf{Single-task:}
For the single-task models, the best performing
model on the entire SOC dataset ($S_{all}$ in \tabref{tab:Attribute}) is NLDF~\cite{Luo2017CVPR} ($M_{S}=0.818$),
followed by RFCN~\cite{wang2016saliency} ($M_{S}=0.814$).
MDF~\cite{li2015visual} and AMU~\cite{zhang2017amulet}
use edge cues to promote the saliency map
but fail to achieve the ideal goal.
Aiming at using the local region information of
images, MC~\cite{zhao2015saliency}, MDF~\cite{li2015visual},
ELD~\cite{lee2016deep}, and DISC~\cite{chen2016disc} try
to use superpixel methods to segment images into regions
and then extract features from these regions, which is complex and time-consuming.
To further
improve the performance, UCF~\cite{zhang2017learning},
DSS~\cite{HouPami18Dss}, NLDF~\cite{Luo2017CVPR}, and
AMU~\cite{zhang2017amulet} utilize the FCN to improve the performance of SOD ($S_{sal}$ in \tabref{tab:Atr_Performance}).
Some other methods such as DCL~\cite{li2016deep} and IMC~\cite{zhang2017deep} try to combine superpixels with FCN to build a powerful model.
Furthermore, RFCN~\cite{wang2016saliency} combines two
related cues including edges and superpixels into FCN to obtain the good performance ($M_{F}=0.435$, $M_{S}=0.814$) over the overall dataset.

\vspace{5pt}
\textbf{Multi-task:}
Different from models mentioned above, MSR~\cite{li2017instance} detects
the instance-level salient objects using three closely related steps:
estimating saliency maps, detecting salient object contours, and
identifying salient object instances.
It creates a multi-scale saliency refinement network that
results in the highest performance ($S_{all}$).
Other two multi-task models DS~\cite{li2016deepsaliency} and
WSS~\cite{wang2017learning} utilize the segmentation and
classification results simultaneously to generate the saliency maps,
obtaining a moderate performance. 
It is worth mentioning that although WSS is a weakly supervised multi-task model, it still achieves comparable performance to other
single-task, fully supervised models. So, the weakly-supervised and
multi-task based models can be promising future directions.

\subsection{Attributes-based Evaluation}\label{sec:Attributes_evaluation}
We assign the salient images with attributes as discussed in
\secref{sec:our_dataset} and \tabref{tab:Attr}. Each attribute
stands for a challenging problem faced in the real-world scenes.
The attributes allow us to identify groups of images with a
dominant feature (\eg, presence of clutter),
which is crucial to illustrate the performance of SOD models and
to relate SOD to application-oriented tasks.
For example, sketch2photo application~\cite{chen2009sketch2photo}
prefers models with good performance on big objects, which can be identified by attributes-based performance evaluation methods.

\textbf{Results.} In \tabref{tab:Atr_Performance},
we show the performance on subsets of
our dataset characterized by a particular attribute.
Due to space limitation, in the following parts, we only select some representative attributes for further analysis.
More details can be found in the supplementary material.

\emph{Big Object} (BO) scenes often occur when objects are in a close distance with the camera,
in which circumstances the tiny text or patterns would always be seen clearly.
In this case, the models which prefer to focus on local information will be mislead seriously, leading to a considerable (\eg, 28.9\% loss for DSS~\cite{HouPami18Dss},
20.8\% loss for MC~\cite{zhao2015saliency} and 23.8\% loss for RFCN~\cite{wang2016saliency}) loss of performance.

\begin{table*}[t!]
  \centering
  \scriptsize
  \renewcommand{\arraystretch}{1}
  \renewcommand{\tabcolsep}{0.2mm}
  \caption{Attributes-based performance on our SOC salient objects sub-dataset.
  For each model, the score corresponds to the average structure
  similarity $M_{S}$ (in \secref{sec:metrics}) over all datasets with that specific attribute (\eg, CL).
  The higher the score the better the performance.
  \textbf{Bold} for the best.
  The average salient-object performance $S_{sal}$ is presented in the first row using the structure similarity $S$.
  The symbol of $^{+}$ and $^{-}$ indicates \emph{increase} and \emph{decrease} compared to the average ($S_{sal}$) result, respectively}\label{tab:Atr_Performance}

  \begin{tabular}{clllllllllllllllll}
  \toprule
  \multicolumn{1}{c}{} & \multicolumn{13}{c}{Single-task} & \multicolumn{3}{c}{Multi-task} \\

  \cmidrule(l){1-1}      \cmidrule(l){2-14}              \cmidrule(l){15-17}
  \rowcolor{red!10}
  \textbf{\emph{Attr}} & LEGS & MC & MDF & DCL & AMU & RFCN & DHS & ELD & DISC & IMC & UCF &
  DSS & NLDF & DS & WSS & MSR\\ 
  & \cite{wang2015deep} & \cite{zhao2015saliency} & \cite{li2015visual} & \cite{li2016deep} & \cite{li2016deep} & \cite{wang2016saliency} & \cite{liu2016dhsnet} & \cite{lee2016deep} & \cite{chen2016disc} & \cite{zhang2017deep} & \cite{zhang2017learning} & \cite{HouPami18Dss} & \cite{Luo2017CVPR} & \cite{li2016deepsaliency} & \cite{wang2017learning} & \cite{li2017instance} \\ 
  \cmidrule(l){1-1}      \cmidrule(l){2-14}              \cmidrule(l){15-17}

  $S_{sal}$ & .607 & .619 & .610 & .705 & .705 & .709 & \textbf{.728} & .664 & .629 & .679 & .678 &.698 & .714 & .719 & .676 & \textbf{.748}\\
  \midrule
  \textbf{\emph{AC}} & .625 & .631& .614 & .734 & .736 & .744 & \textbf{.745} & .673 & .644 & .702 & .714 & .726 & .737 & .764 & .691 & \textbf{.789}\\
  \textbf{\emph{BO}} & .509 & .490& .461$^{-}$ & .610 & .569 & .540 & .590 & .576 & .517 & \textbf{.701}$^{+}$ & .636 & .496$^{-}$ & .568 & .685 & .566 &.667\\
  \textbf{\emph{CL}} & .620 & .635& .566 & .699 &.708 & .714 & \textbf{.743} & .658 &  .635 & .696 & .704 &.677$^{-}$ & .713 & .729 & .678 &\textbf{.756}\\
  \textbf{\emph{HO}} & .666 & .666& .648 & .745 & .755 & .759 & \textbf{.766} & .706 &.681& .715 & .744 & .748 & .755 & .756 & .707 & \textbf{.777}\\
  \textbf{\emph{MB}}& .543$^{-}$ & .603& .615 & .693 & .706 & .715 & \textbf{.722} & .639 & .600 & .689 & .682 &.695 & .685 & .711 & .641 & \textbf{.757}\\
  \textbf{\emph{OC}} & .609 & .617& .608 & .708$^{+}$ & \textbf{.725}$^{+}$ & .711 &.716 & .658 & .630 & .672 & .701$^{+}$&.689 & .709 & .725$^{+}$ & .672 & \textbf{.740}\\
  \textbf{\emph{OV}} & .548 & .584& .568 & .699 & \textbf{.708}$^{+}$ & .687 &.706 & .637 & .573 &.693$^{+}$ & .685$^{+}$&.665 & .688 & .722$^{+}$ & .624 &\textbf{.743}\\
  \textbf{\emph{SC}} & .608 & .620& .669$^{+}$ & .738 & .731 & .735 & \textbf{.763}& .688 &  .653 & .690 & .722$^{+}$ &.746$^{+}$ & .745 & .724 & .677 & \textbf{.773}\\
  \textbf{\emph{SO}} & .573$^{-}$& .601 & .621 & .691& .685 & .698 & \textbf{.713} & .644 &  .614 & .648$^{-}$ & .650 &.696$^{-}$ & .703 & .696 & .659 & \textbf{.730}\\
  \bottomrule
  \end{tabular}
\end{table*}

However, the performance of IMC~\cite{zhang2017deep} model goes up for a slight margin of 3.2\% instead. After taking a deeper look of the pipeline of this model, we came up a reasonable explanation.
IMC uses a coarse predicted map to express semantics and utilizes over-segmented images to supplement the structural information, achieving a satisfying result on type \emph{BO}.
However, over-segmented images cannot make up the missing details, causing 4.6\% degradation of performance on the type of \emph{SO}.

\emph{Small Object} (SO) is tricky for all SOD models.
All models encounter performance degradation (\eg, from DSS~\cite{HouPami18Dss} -0.3\% to LEGS~\cite{wang2015deep} -5.6\%),
because \emph{SOs} are easily ignored during down-sampling of CNNs.
DSS~\cite{HouPami18Dss} is the only model that has a slight decrease of performance on type \emph{SO}, while it has
the biggest (28.9\%) loss of performance on type \emph{BO}.
MDF~\cite{li2015visual} uses multi-scale superpixels as the input of network, so it retains the details of small objects well.
However, due to the limited size of superpixels, MDF can not efficiently sense the global semantics, causing a big failure on type \emph{BO}.

\emph{Occlusions} (OC) scenes in which objects are partly obscured.
Thus, it requires SOD models to capture global semantics to make up for the incomplete information of objects.
To do so, DS~\cite{li2016deepsaliency} \& AMU~\cite{zhang2017amulet} made use of the
multi-scale features in the down-sample progress to generate a fused saliency map;
UCF~\cite{zhang2017learning} proposed an uncertain learning mechanism to learn uncertain convolutional features.
All these methods try to get saliency maps containing both global and local features.
Unsurprisingly, these methods have achieved pretty good results on type \emph{OC}.
Based on the above analyses, we also find that these three models perform very well on the scenes requiring
more semantic information like type \emph{AC}, \emph{OV} and \emph{CL}.

\emph{Heterogeneous Object} (HO)
is a common attribute in nature scenes.
The performance of different models on type \emph{HO} gets some improvement to their average performances respectively, all fluctuating from
3.9\% to 9.7\%.
We suspect this is because type \emph{HO} accounts for a significant proportion of all datasets, objectively making models more fitting to this attribute.
This result in some degree confirms our statistics in \figref{fig:Attributes_distribution}.

\section{Discussion and Conclusion}
To our best knowledge, this work presents the currently largest scale performance
evaluation of CNNs based salient object detection models.
Our analysis points out a serious \emph{data selection bias} in existing SOD datasets.
This design bias has lead to state-of-the-art SOD algorithms almost achieve saturated high
performance when evaluated on existing datasets, but are still far from being satisfactory
when applied to real-world daily scenes.
Based on our analysis, we first identify 7 important aspects that a comprehensive and
balanced dataset should fulfill.
We firstly introduces a high quality SOD dataset, \textbf{SOC}.
It contains salient objects from daily life in their natural environments
which reaches closer to realistic settings. The SOC dataset will evolve and grow over time and will enable research possibilities in
multiple directions, \eg, salient object subitizing~\cite{zhang2015salient},
instance level salient object detection~\cite{li2017instance},
weakly supervised based salient object detection~\cite{wang2017learning}, \emph{etc}.
Then, a set of attributes (\eg, \emph{Appearance Change}) is proposed in the attempt
to obtain a deeper insight into the SOD problem, investigate the pros and cons of the SOD algorithms, and objectively assess the model performances over different
perspectives/requirements.
Finally, we report attribute-based performance assessment on our SOC dataset.
The results open up promising future directions for model development and comparison.



%
%
\bibliographystyle{splncs04}
\bibliography{SOCBenchmark}

\end{document}